%% file: acl_latex.tex
\pdfoutput=1

\documentclass[11pt]{article}

\usepackage[preprint]{acl} 

\usepackage{times}
\usepackage{latexsym}

\usepackage[T1]{fontenc}

\usepackage[utf8]{inputenc}

\usepackage{microtype}

\usepackage{inconsolata}

\usepackage{graphicx}

\usepackage{enumitem}

\usepackage{booktabs}
\usepackage{gb4e}
\noautomath 
\usepackage{xcolor}
\usepackage{tipa}

\title{Massively Multilingual Joint Segmentation and Glossing}

\author{Michael Ginn${ }^{1}$ \quad Lindia Tjuatja${ }^{2}$ \quad  Enora Rice${ }^{1}$ \quad  Ali Marashian${ }^{1}$ \\ {\bf Maria Valentini}${ }^{1}$ \quad {\bf Jasmine Xu}${ }^{2}$ \quad {\bf Graham Neubig}${ }^{2}$ \quad {\bf Alexis Palmer}${ }^{1}$\\
    ${ }^{1}$University of Colorado Boulder \quad ${ }^{2}$Carnegie Mellon University  \\ \texttt{michael.ginn@colorado.edu} \quad \texttt{alexis.palmer@colorado.edu} }

\begin{document}
\maketitle
\begin{abstract}
Automated interlinear gloss prediction with neural networks is a promising approach to accelerate language documentation efforts. However, while state-of-the-art models like \textsc{GlossLM} \citep{ginn-etal-2024-glosslm} achieve high scores on glossing benchmarks, user studies with linguists have found critical barriers to the usefulness of such models in real-world scenarios \citep{rice-etal-2025-interdisciplinary}. In particular, existing models typically generate morpheme-level glosses but assign them to whole words without predicting the actual morpheme boundaries, making the predictions less interpretable and thus untrustworthy to human annotators.


We conduct the first study on neural models that \textbf{jointly predict interlinear glosses and the corresponding morphological segmentation} from raw text. We run experiments to determine the optimal way to train models that balance segmentation and glossing accuracy, as well as the alignment between the two tasks. We extend the training corpus of \textsc{GlossLM} and pretrain \textsc{PolyGloss}, a family of seq2seq multilingual models for joint segmentation and glossing that outperforms \textsc{GlossLM} on glossing and beats various open-source LLMs on segmentation, glossing, and alignment. In addition, we demonstrate that \textsc{PolyGloss} can be quickly adapted to a new dataset via low-rank adaptation.


\end{abstract}

\section{Introduction}
\begin{figure}
    \centering
    \includegraphics[width=0.9\linewidth]{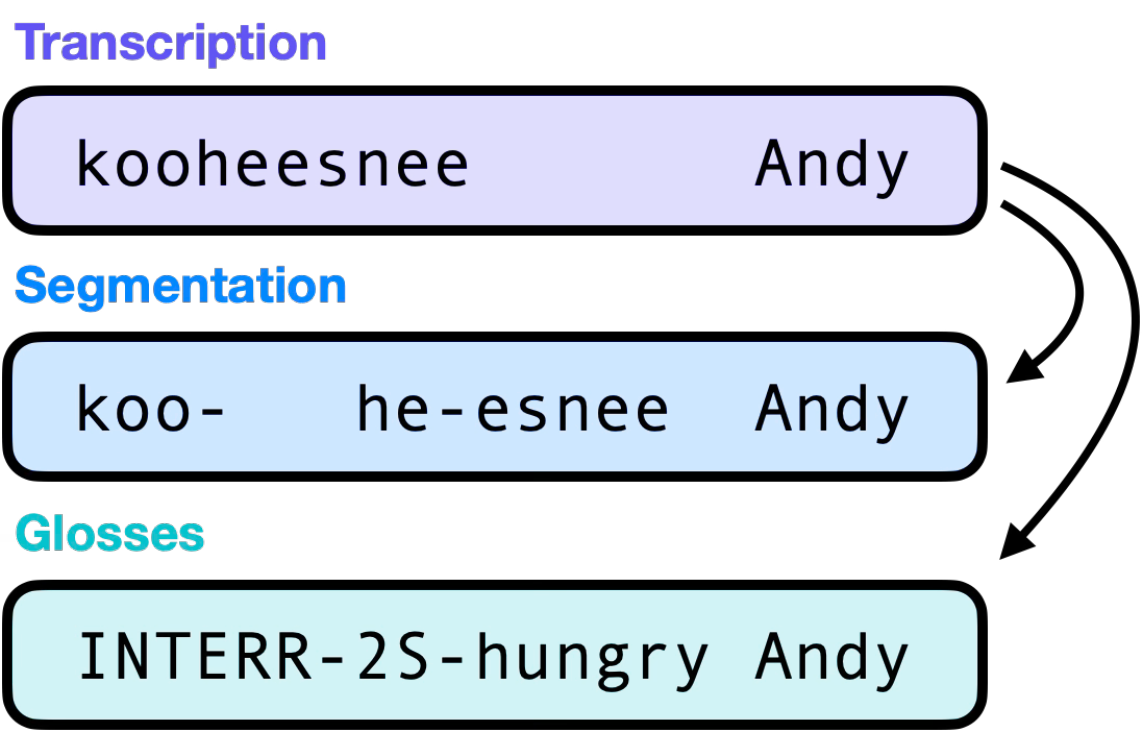}
    \caption{An interlinear glossed text example, showing the Arapaho for \textit{"Are you hungry, Andy?"}. Our model predicts the segmentation and gloss line from the transcribed text.}
    \label{fig:overview}
\end{figure}

Nearly half of the world's 7,000 languages face extinction. For many speakers and linguists of these languages, \textbf{language documentation} has become an urgent goal. Documentation projects commonly involve the creation of interlinear glossed text (IGT), a dense annotation format combining morphological segmentation, tagging, and translation (\autoref{fig:overview}). Due to its structured format and common usage among linguists, IGT has proven useful for linguistic analysis \citep{bender-etal-2013-towards, zamaraeva-2016-inferring, moeller-etal-2020-igt2p}, language pedagogy \citep{doi:10.1177/13621688211020423, Bonilla_Carvajal_2025}, and development of language technology such as taggers \citep{Georgi2016FromAT}, searchable text databases \citep{blokland2019using, rijhwani-etal-2023-user}, educational tools \citep{uibo_building_2017,chaudhary-etal-2023-teacher}, and machine translation systems \citep{zhou2020usinginterlinearglossespivot, ramos-etal-2025-grammamt}.

Creating IGT is expensive, and a number of studies have proposed methods to automate IGT production with statistical and neural methods \citep{mcmillan-major-2020-automating,zhao-etal-2020-automatic,ginn-etal-2024-teach}. In all of these studies, including the 2023 SIGMORPHON shared task \citep{ginn-etal-2023-findings}, the task is formulated as predicting the gloss line from the transcription or segmentation line. The former is more difficult (but also more useful), as it requires the model to infer morphological segmentation in addition to predicting glosses, and has been the primary focus of recent work.

Though state-of-the-art glossing models such as \textsc{GlossLM} \citep{ginn-etal-2024-glosslm} have achieved high accuracy across many languages, \citet{rice-etal-2025-interdisciplinary} discovered several issues when using these models in a realistic documentation scenario: 
\begin{enumerate}
    \item First, documentary linguists typically perform \textbf{explicit morphological segmentation} before glossing each morpheme, so a model that produces glosses directly---without exposing the implicit segmentation---is confusing, less interpretable, and difficult to trust. \label{problem1}
    \item Second, the model produced very \textbf{inaccurate glosses} in two of the three languages studied, with the participants agreeing that correcting the predicted outputs would be more difficult than annotating from scratch, or using a simpler lookup-based method. \label{problem2}
    \item Third, the model often predicted gloss labels which were unfamiliar or unlike the glossing conventions preferred by the participants, and the existing system provided no way to \textbf{adapt its labels to the preferred conventions}.  \label{problem3}
\end{enumerate}

\noindent In this work, we address these three concerns, building on the approach of \textsc{GlossLM}. We release an improved version of the \textsc{GlossLM} corpus, which adds 91k examples for a total of 341k examples, improves standardization and formatting, and ensures alignment between morphological segmentation and glosses. We train a multilingual model on the dataset for \textbf{joint segmentation and glossing}, optimizing for performance on both tasks, as well as \textbf{alignment between the two tasks}, outperforming various small LLMs and satisfying \autoref{problem1}. We show that per-language perplexity can roughly predict glossing accuracy for any language, addressing \autoref{problem2} by enabling an automatic glossing system to avoid giving low-quality predictions, or to fall back to a simpler model. Finally, we show that \textsc{PolyGloss} can be rapidly adapted to small labeled datasets via low-rank adaptation, satisfying \autoref{problem3}. Unlike prior work that trains monolingual glossing models, we focus on creating \textbf{a single multilingual model that can be used out-of-the-box on many languages.} Our models and dataset are available on HuggingFace\footnote{\url{https://huggingface.co/collections/lecslab/polygloss}} and our code is on GitHub.\footnote{\url{https://github.com/lecs-lab/polygloss}}

\section{Corpus}
\begin{table}[h]
\small
    \centering
    \begin{tabular}{l|r}
        \toprule
        Statistic & Count \\
        \midrule
        Total examples & 353,266 \\
        Unique languages & 2,077 \\
        Train examples & 340,251 \\
        Eval examples & 6,148 \\
        Test examples & 6,867 \\
        \midrule
        No glottocode & 13,428 \\
        No metalang. glottocode & 10,894 \\
        No segmentation & 93,648 \\
        No translation & 5,921 \\
        Misaligned & 34,894 \\
        \bottomrule
    \end{tabular}
    \caption{\textsc{PolyGloss} corpus statistics}
    \label{tab:dataset_stats}
\end{table}
\noindent We create an enhanced version of the \textsc{GlossLM} corpus with significantly improved formatting. We consistently handle punctuation across all sources, ensuring that sentence-ending punctuation is surrounded by spaces while gloss-internal punctuation remains unchanged, as in the following Tsez example:

\begin{small}
\begin{exe}
  \ex
  \label{ex:dido_gloss}
  \gll Žeda kidbeqor kurno lel yayrno . \\
       žeda-a kid-qor kur-n lel y-ayr-n \\
       DEM1.IIPL.OBL-ERG girl-POSS.LAT throw-PFV.CVB wing II-lead-PST.UNW 
\end{exe}
\end{small}
We fixed a number of source-specific formatting issues. For example, we noticed 4,882 instances in the Arapaho data where ",." was used inside glosses, and we confirmed with the original annotator that this was an error. Across sources, we identified instances where the morphological segmentation was \textbf{misaligned} with the glosses---that is, cases where there was a mismatch in either the number of words or the number of segments within a word (see \S\ref{sec:alignment_score} for more discussion of alignment). In these cases, if the segmentation field does not include any segmentation markers (and the gloss field does), we set the segmentation to blank. Otherwise, we keep the segmentation, but ensure the offending examples are within the training split, so as not to affect evaluation. 

We also incorporate additional IGT data into the original \textsc{GlossLM} corpus. The largest of these is the Fieldwork dataset \citep{he-etal-2024-wav2gloss}, which collects 80,461 IGT instances for 37 languages. We also update the IMTVault dataset \citep{nordhoff-kramer-2022-imtvault} to the newest version (1.2), which includes IGT scraped from new linguistic publications, adding 39,741 examples. After removing 20,116 duplicates and filtering very low-quality examples,\footnote{Examples which only contained punctuation or whitespace} we have \textbf{91,416 new unique examples} compared to the original \textsc{GlossLM} corpus. We introduced an auditing process on our dataset to quantify issues, and report full statistics in \autoref{tab:dataset_stats}. We add two new languages as evaluation languages from the Fieldwork dataset: Hokkaido Ainu (\texttt{ainu1240}) and Ruuli (\texttt{ruul1235}). Our dataset splits for all evaluation languages are reported in \autoref{tab:lang_splits}.

\begin{table}[h]
    \small
    \centering
    \begin{tabular}{l | c c c}
        \toprule
        Language & Train & Eval & Test \\
        \midrule
        Arapaho (\texttt{arp}) & 36776 & 4687 & 4499 \\
        Tsez (\texttt{ddo}) & 3626 & 444 & 442 \\
        Gitksan (\texttt{git}) & 89 & 42 & 37 \\
        Uspanteko (\texttt{usp}) & 8338 & 170 & 566 \\
        Ainu (\texttt{ain}) & 6726 & 218 & 590 \\
        Lezgi (\texttt{lez}) & 646 & 51 & 53 \\
        Natugu (\texttt{ntu}) & 786 & 99 & 99 \\
        Nyangbo (\texttt{nyb}) & 1221 & 225 & 248 \\
        Ruuli (\texttt{ruc}) & 2158 & 212 & 333 \\
        \bottomrule
    \end{tabular}
    \caption{Number of examples for each evaluation language across train, eval, and test splits.}
    \label{tab:lang_splits}
\end{table}

\section{Evaluation}
\label{sec:eval}
Since our model performs joint glossing and segmentation, we compute metrics for both, as well as an alignment score (\S \ref{sec:alignment_score}) between the two.

\subsection{Glossing}
We compute a number of metrics for gloss prediction. Departing from prior work, we use \textbf{morpheme error rate} (c.f. word error rate as used in speech recognition) as our primary metric. While prior work \citep{mcmillan-major-2020-automating, ginn-etal-2023-findings} used morpheme-level accuracy as the primary metric, this assumes that the output has the correct number of morphological glosses. If there is a gloss inserted or deleted, all subsequent glosses will be counted as incorrect. Instead, we compute the morpheme error rate by first inserting \texttt{[SEP]} tokens between the glosses for each word and computing the edit distance, normalized to the length of the gold label sequence. The range is 0 or greater; a score higher than 1 is possible if the predicted sequence is longer than the gold sequence. In addition, we compute word and character error rates, BLEU scores (at all three levels of granularity), and morpheme and word-level accuracy.

\subsection{Segmentation}
We use standard metrics for evaluating segmentation. We primarily report the modified F1 score as defined in \citet{mager-etal-2020-tackling}, which computes precision based on morphemes in the predicted segmentation also occurring in the gold label, and vice versa for recall. We also compute character-level edit distance and whole-word accuracy. 

\subsection{Alignment}
\label{sec:alignment_score}
A key goal in this study is to predict  morphological segmentations and glosses that are aligned with one another, making the gloss predictions more interpretable and trustworthy for a human annotator. To measure this, we propose a novel \textbf{alignment score}, which is computed based on predictions with no reference to the gold sequence.\footnote{That is, a model could achieve a perfect alignment score while predicting incorrect glosses and segmentation.} First, the segmentation and gloss predictions are converted into \textit{abstract sequences} that represent morphological structure. Ignoring punctuation, each morpheme sequence is converted to a single ``x'' character, and morpheme boundaries (``-'' and ``='') are left unchanged, as in the following example:

\begin{center}
    \texttt{the cat-s ru-n} $\Rightarrow$ \texttt{x x-x x-x} \\
    \texttt{DET cat-PL run.SG} $\Rightarrow$ \texttt{x x-x x}
\end{center}

\noindent Then, the character-level edit distance is computed between the abstracted gloss and segmentation sequences, ranging from 0 to infinity. We normalize the edit distance by the length of the longer sequence,\footnote{Unlike with the standard error rate, we don't know which sequence is correct if there is a mismatch.} and subtract from 1 to give a score in $[0,1]$ where 1 is a perfect score. In this example, the alignment score is 0.78. 

\section{Model}
\subsection{Task Format}
Using the \textsc{PolyGloss} corpus, we perform continued pretraining on a pretrained multilingual LLM for both segmentation and glossing. We train the model to predict glosses from both the segmented and unsegmented transcription, but we only evaluate on the latter, as it is the more difficult and realistic setting. We experiment with three different approaches for combining the two tasks and report results in \ref{sec:ablation2}.

\paragraph{Multitask Prediction} In this setting, separate training examples are created for segmentation and for glossing. The examples are formatted as in the following Vera'a language example (replacing "glosses" with "segmentation" when appropriate):

\vspace{3pt}
\noindent\fbox{\begin{minipage}{\linewidth}\small \texttt
    Predict the glosses for the following text in \textcolor{blue}{Vera'a}. \\
    Text in \textcolor{blue}{Vera'a}: \textcolor{blue}{o wōlēn 'ēqēk} \\
    Translation in \textcolor{blue}{English}: \textcolor{blue}{Oh, over there is my garden} \\
    Glosses: \textcolor{blue}{INTERJ you.know-ZERO=ART garden-1SG}
\end{minipage}}
\vspace{3pt}

\noindent This setting is simple and allows for simultaneous inference of both glosses and segmentation. However, there is greater risk of misalignment, since the two tasks are trained separately and alignment is not enforced.

\paragraph{Concatenated Prediction} Morphological segmentation and interlinear glosses are not in fact distinct tasks, as the latter depends inherently on the former. In this setting, we model this dependency by training the model to predict the segmentation followed by the glosses:

\vspace{3pt}
\noindent\fbox{\begin{minipage}{\linewidth}\small \texttt
    Predict the morphological segmentation and glosses for the following text in \textcolor{blue}{Vera'a}. \\
    Text in \textcolor{blue}{Vera'a}: \textcolor{blue}{o wōlēn 'ēqēk} \\
    Translation in \textcolor{blue}{English}: \textcolor{blue}{Oh, over there is my garden} \\
    Segmentation: \textcolor{blue}{o wōlē-0=n 'ēqē-k} \\
    Glosses: \textcolor{blue}{INTERJ you.know-ZERO=ART garden-1SG}
\end{minipage}}
\vspace{3pt}

\noindent This introduces a natural dependency due to the causal training objective: while generating the gloss string, the model can attend to tokens in the segmentation. Of course, this is not a strict constraint, and carries the risk that a bad segmentation will affect the glosses as well. Since not all training examples have segmentation labels, we also create glossing examples in the multitask style shown in the previous section.

\paragraph{Interleaved Prediction} While the concatenated setting trains the model with an implicit relationship between segments and glosses, it is still possible to generate misaligned predictions. In this setting, we introduce a hard constraint using an interleaved format that explicitly aligns segments and glosses. In this format, each gloss label is immediately followed by the corresponding morpheme in parentheses.

\vspace{3pt}
\noindent\fbox{\begin{minipage}{\linewidth}\small \texttt
    Predict the glosses and morphological segmentation (in parentheses) for the following text in \textcolor{blue}{Vera'a}. \\
    Text in \textcolor{blue}{Vera'a}: \textcolor{blue}{o wōlēn 'ēqēk} \\
    Translation in \textcolor{blue}{English}: \textcolor{blue}{Oh, over there is my garden} \\
    Output: \textcolor{blue}{INTERJ(o) you.know(wōlē)-ZERO(0)=ART(n) garden('ēqē)-1SG(k)}
\end{minipage}}
\vspace{3pt}

\noindent We hypothesized that this setting would have the best alignment, as any well-formed output should be perfectly aligned.

\subsection{Base Model}
 Following \citep{ginn-etal-2024-glosslm}, we perform continued pretraining on ByT5 \citep{xue-etal-2022-byt5}, a byte-level encoder-decoder transformer language model based on the T5 architecture. By using byte-level tokenization, ByT5 avoids the issues that arise for rare languages with subword tokenizers, and has been shown to outperform T5 on multilingual glossing \citep{he-etal-2023-sigmorefun}. We also experimented with finetuning an instruction-tuned decoder LLM, using Qwen 3 0.6B \citep{yang2025qwen3technicalreport}, known to be a strong multilingual model. However, we saw poor results, discussed in \autoref{sec:base_model_appendix}. We train one ByT5-based models for each task format in the preceding section, using the \texttt{byt5-base} checkpoint with 580M parameters.





\subsection{Training} Due to the high cost of training runs, we do not perform extensive tuning. We train all models in bf16, using the AdamW optimizer with default parameters, a linear learning rate warmup for the first 3\% of steps, cosine learning rate decay, and gradient clipping with max norm of 1. We train all models using 4 GH200 GPUs. For the ByT5-based models, we used a learning rate of 5E-5, batch size of 64, and 15 epochs. Evaluation uses beam search with default parameters and 2 beams. Parameters for the Qwen-based model are given in \autoref{sec:base_model_appendix}.

\section{Baselines}
\subsection{Multilingual Baselines}
We compare our best model to the following multilingual models for glossing:

\paragraph{\textsc{GlossLM}} We compare glossing performance to the pretrained \textsc{GlossLM} model, primarily to ensure that incorporating segmentation does not cause glossing performance to regress. As this model was not trained for segmentation, we cannot compare across all metrics. Additionally, the original \textsc{GlossLM} model did not include all of the segmented training data, so this is not a perfect head-to-head comparison. 

\paragraph{In-Context Learning} Following \citet{ginn-etal-2024-teach}, we use LLMs to predict both the segmentation and glosses in a single pass using the interleaved format. We retrieve ten similar examples (via chrF score) from the training set to provide in-context. We test three models: the \textbf{Qwen 3 0.6B model} with thinking \citep{yang2025qwen3technicalreport} and the \textbf{Cohere Aya Expanse 8B model} \citep{dang2024ayaexpansecombiningresearch} and \textbf{Google Gemma 3 4B model} \citep{gemma_2025} without thinking. These enable a direct comparison between our finetuned seq2seq models and ICL with LLMs of similar size.

\subsection{Monolingual Baselines}
\label{sec:monolingual_baselines}
In addition, we run preliminary experiments using monolingual models to understand the effect of different approaches to joint training and monolingual versus multilingual training. We use the following baselines, and provide comparisons in \ref{sec:ablation1} and \ref{sec:ablation2}:

\paragraph{Finetuned ByT5 (separate)} For each language in our test set, we train monolingual models on ByT5 for segmentation and glossing in the Multitask format, using \textbf{separate models for each task}. We use the same hyperparameters as our pretrained model, except we set a max 30 epochs and use early stopping with patience 5 to prevent overfitting. We do not perform extensive tuning.

\paragraph{Finetuned ByT5 (joint)} In addition, we train monolingual models for \textbf{joint segmentation and glossing} using the Interleaved format and the same hyperparameters as the prior baseline.

\paragraph{Pipeline} We train monolingual pipelines of two ByT5 models, where the first model predicts the morphological segmentation and the second model predicts glosses given segments. In this baseline, there is clear risk of error propagation, as incorrect segmentation likely results in incorrect glosses.

\paragraph{Hard Attention Transformer} Following \citet{girrbach-2023-tu}, we train monolingual hard attention transformers using straight-through gradient estimation. The model is only trained on glosses, but generates an explicit latent segmentation based on the hard attention between glosses and characters in the input.

\section{Results}

\input{latex/tables/mer_glossing_multiling}
\input{latex/tables/f1_segmentation_multiling}
\input{latex/tables/alignment_multiling}

We compare our best \textsc{PolyGloss} model (ByT5 architecture and Interleaved format) with multilingual baselines on the test set for glossing (\autoref{tab:mer-glossing}), segmentation (\autoref{tab:mer-segmentation}), and alignment (\autoref{tab:alignment}). Results for all other metrics, such as BLEU score, morpheme accuracy, and word-level scores are available on our GitHub.

Overall, the multilingual \textsc{PolyGloss} model is state-of-the-art on glossing and very strong on segmentation. It outperforms \textsc{GlossLM} for glossing on the three overlapping evaluation languages, likely due to both better dataset preprocessing and improved training hyperparameters. It also far outperforms in-context learning approaches with LLMs that are orders of magnitude larger (220M vs 0.6B-8B). The smallest LLM (Qwen 0.6B) often struggles to conform to the desired format, resulting in misaligned outputs. The larger models (Gemma 4B and Aya 8B) have much higher alignment scores, but still struggle to perform glossing or segmentation accurately. More details on specific types of failures are described in \autoref{sec:appendix_llm_icl}. While significantly larger LLMs might show better results, our model is clearly state-of-the-art given constraints on both model size and training budget. 

\subsection{Predicting Performance on Other Languages}
\begin{figure}[h]
    \centering
    \includegraphics[width=\linewidth]{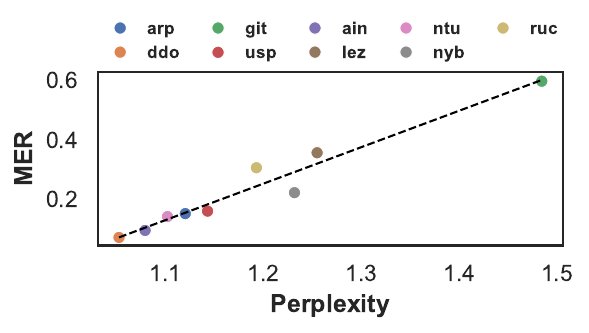}
    \caption{Relationship between validation set perplexity for a given language and glossing performance, as measured by morpheme error rate. There is a strong correlation ($r^2=0.951$), indicating that perplexity can be used as a rough predictor of glossing performance.}
    \label{fig:correlation}
\end{figure}
\noindent As identified in \citet{rice-etal-2025-interdisciplinary}, an issue with \textsc{GlossLM} was that there was no good way to predict glossing performance on a given language that is not one of our nine selected evaluation languages. For the \textsc{PolyGloss} model (ByT5, Interleaved), we compute per-language perplexity on our validation dataset and demonstrate that it has a strong correlation ($r^2=0.951$) with our target metric (\autoref{fig:correlation}).

This provides a practical heuristic for the use of our model in real-world settings. A glossing software such as Plaid\footnote{\url{https://www.langdoc.net/t/introducing-plaid/1250}} could set an acceptable error rate threshold, and use the \textsc{PolyGloss} model to predict glosses only if the language's expected error rate is below that threshold. If not, then it will likely be a better user experience to either fall back to a simple method (such as predicting the highest-frequency gloss) or not showing predictions at all.

\section{Ablations}
Per-language breakdowns are provided for all ablations in \autoref{sec:ablation_numbers}.
\subsection{Effect of Joint Training}
\label{sec:ablation1}
\begin{figure}[tbh]
    \centering
    \includegraphics[width=\linewidth]{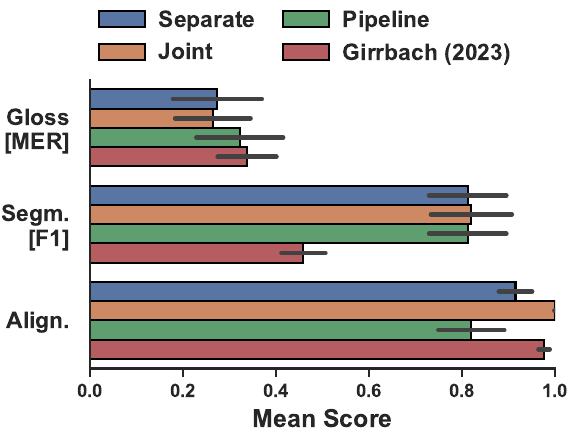}
    \caption{Scores for monolingual models using various approaches to multitask training. A lower glossing MER is better; higher is better for the other two metrics. Scores are averaged across nine languages and reported with standard error.}
    \label{fig:ablation1}
\end{figure}
The best \textsc{PolyGloss} model is trained jointly on segmentation and glosses in a single training example. To disentangle the effect of this joint training approach, we compare three monolingual ByT5-based approaches: training for glossing and segmentation separately, jointly, and in a model pipeline (see \autoref{sec:monolingual_baselines}). We also evaluate the \citet{girrbach-2023-tu} approach, which induces a latent segmentation using hard attention. We report the average scores in \autoref{fig:ablation1}.\footnote{Error bars are large because of the variance across languages, and should not be used to determine significance.} The \textbf{Joint} setting is superior on all three metrics (and near-perfect on alignment), suggesting a harmonious relationship between the two training tasks. While the \textbf{Separate} setting is similar to the former on glossing and segmentation, its alignment score is significantly worse. This is unsurprising, as the two separate models for glossing and alignment are not guaranteed to produce the same errors, and thus may generate misaligned outputs for the same input. For example, the separated models make the following poorly aligned prediction for a Nyangbo sentence (alignment score of 0.882):

\begin{small}
\begin{exe}
  \ex
  \label{ex:bad_sep}
  \gll vũn\textipa{\textopeno} gagãlĩ g\textipa{\:E} enu budzyu\textipa{\:d}í y\textipa{\:E} \\
       vũn\textipa{\textopeno} gagãlĩ g\textipa{\:E} e-nu bu-dzyu\textipa{\:d}í y\textipa{\:E} \\
       beverage well-well REL 3SG-be CM-strength 3SG
\end{exe}
\end{small}

\noindent The misalignment occurs in the second word, which is predicted to be a single morpheme but two glosses. Meanwhile, the interleaved joint model predicts the perfectly aligned glosses and morphemes (which are also more accurate):

\begin{small}
\begin{exe}
  \ex
  \label{ex:good_sep}
  \gll vũn\textipa{\textopeno} gagãlĩ g\textipa{\:E} enu budzyu\textipa{\:d}í y\textipa{\:E} \\
       vũn\textipa{\textopeno} gagãlĩ g\textipa{\:E} e-nu bu-dzyu\textipa{\:d}í y\textipa{\:E} \\
       greet be\_hard REL 3SG-be CM-dawadawa\_tree FOC
\end{exe}
\end{small}
\noindent For real-world usage, generating aligned outputs is critical in addition to achieving high accuracy.

The \textbf{Joint} approach is also clearly superior to the \textbf{Pipeline} setting for glossing and alignment (though segmentation is very similar). Because the pipeline is not trained end-to-end, error in the intermediate predictions is catastrophic for the gloss predictions, resulting in far higher error than the joint approach. Finally, the jointly trained model outperforms the hard attention approach of \citet{girrbach-2023-tu} on segmentation, where the learned segmentations are not very accurate to the gold labels.

\subsection{Effect of Multilingual Pretraining}
\label{sec:ablation2}
\begin{figure}[tbh!]
    \centering
    \includegraphics[width=\linewidth]{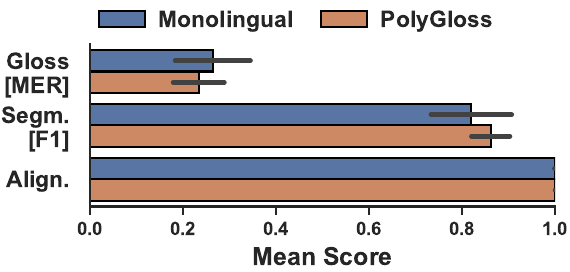}
    \caption{Ablation comparing monolingual joint models (using the interleaved format) and the multilingual \textsc{PolyGloss} using the same format. A lower glossing MER is better; higher is better for the other two metrics. Scores are averaged across nine languages and reported with standard error.}
    \label{fig:ablation2}
\end{figure}

\noindent We compare the monolingual joint models with the \textsc{PolyGloss} multilingual model (\autoref{fig:ablation2}). Both approaches use the interleaved format, enabling a direct comparison. We see that on average, the multilingual model outperforms the monolingual models on glossing and segmentation (and alignment is perfect for both), though the standard error is large. We observed that for the three languages with the most training data (\texttt{arp}, \texttt{usp}, and \texttt{ain}) the monolingual model is slightly superior on glossing. Meanwhile, for all other languages the multilingual model is far superior on glossing, demonstrating clear transfer learning benefits. Furthermore, we believe that serving a single multilingual glossing model is preferable to requiring linguists to train and host their own monolingual models (though we do propose an adaptation method in \autoref{sec:lora}).

\subsection{Effect of Task Format}
\label{sec:ablation3}
\begin{figure}[tbh!]
    \centering
    \includegraphics[width=\linewidth]{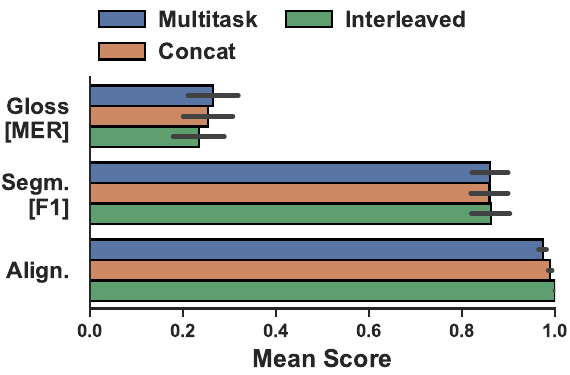}
    \caption{Scores for \textsc{PolyGloss} multilingual models using three different data formats. Scores are averaged across nine languages and reported with standard error.}
    \label{fig:ablation3}
\end{figure}
Results for the three ByT5-based \textsc{PolyGloss} models are very similar regardless of task format (\autoref{fig:ablation3}). The interleaved model is best overall on glossing (average MER 0.234), segmentation (average F1 0.862), and alignment (1.000). The multitask model is slightly worse at glossing, indicating potential benefits from explicitly conditioning the model on the morphological segmentation. As hypothesized, the multitask model struggles the most on alignment, indicating that its glossing and segmentation predictions are accurate but not necessarily aligned. The interleaved format results in perfect alignment thanks to its explicit constraint.

\begin{figure}[tbh!]
    \centering
    \includegraphics[width=\linewidth]{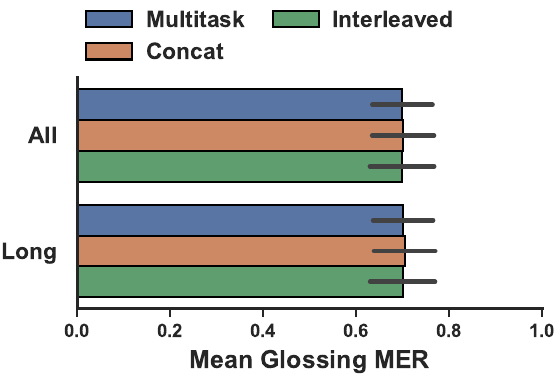}
    \caption{Glossing MER for \textsc{PolyGloss} multilingual models when evaluated either on all data, or only on long inputs (75th percentile in length per language). Full data in \autoref{tab:long-ablation}.}
    \label{fig:ablation5}
\end{figure}

One concern with the interleaved and concatenated formats is that the expected output is roughly twice as long than the joint model, as it includes both the glosses and segmentation. Due to the autoregressive nature of generation, longer sequences can result in degraded accuracy. We test this by filtering the test set to only very long inputs; specifically, we only keep examples where the transcription length is in the 75th percentile for all examples of the same language. We compare glossing results for long inputs between the three formats in \autoref{fig:ablation5}. We observe that average performance is nearly identical on long inputs, thus suggesting there is no concern with the increased length of the interleaved and concatenated formats.

\section{Adapting \textsc{PolyGloss} To New Data}
\label{sec:lora}
Though we strived to collect as much IGT data as possible, it is inevitable that some languages will not be well-represented (or present at all) in the \textsc{PolyGloss} pretraining corpus. Furthermore, as glossing conventions vary between annotators, the glosses produced by \textsc{PolyGloss} may not match the desired schema. \citet{ginn-etal-2024-glosslm} studied full-parameter finetuning for unseen languages, but we argue this is an unrealistic scenario for virtually all documentary linguists (due to both technical difficulty and compute requirements). Instead, we propose the use of low-rank adaptation (LoRA), which drastically reduces the computational cost of training and often requires less data \citep{hu2022lora}. Given a small dataset of new glossed examples, it is feasible to run LoRA adaptation with limited computational resources. For example, a glossing software could periodically train an adapter overnight, enabling predictions in the linguist's target language without any additional effort.

\begin{figure}
    \centering
    \includegraphics[width=1\linewidth]{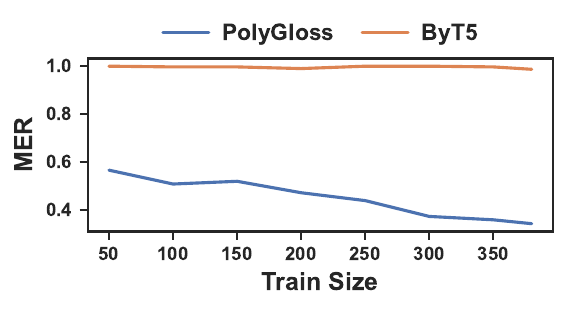}
    \caption{Morpheme error rate for Vamale when training LoRAs on the \textsc{PolyGloss} interleaved model and ByT5 with different size training sets}
    \label{fig:adaptation}
\end{figure}

We simulate a realistic annotation scenario for Vamale (which is never seen during pretraining), training LoRA adapters on increasingly large training datasets (increasing increments of 50 up to the full 380 examples). We use the Vamale data from \citet{yang-etal-2025-linggym} and train rank 8 adapters for 25 epochs, using a batch size of 32; on an A100, the largest training only took 12 minutes. We report MER scores when adapting the interleaved \textsc{PolyGloss} model and a ByT5 base model in \autoref{fig:adaptation}. The \textsc{PolyGloss} quickly adapts to the new language, while the ByT5-based model never improves.

\section{Alignment Score as a Reward Function}
\begin{figure}[b!]
    \centering
    \includegraphics[width=1\linewidth]{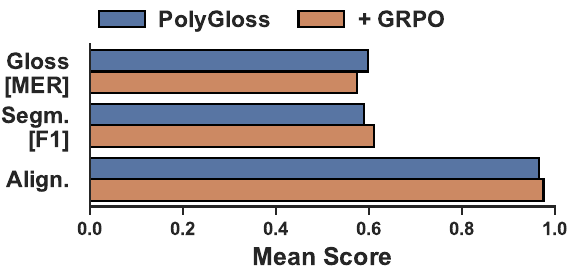}
    \caption{Scores before and after GRPO tuning for Gitksan.}
    \label{fig:grpo}
\end{figure}

Our alignment score can also be used as a scalar reward function for reinforcement learning with verifiable rewards (RLVR, \citealp{deepseekai2025deepseekr1incentivizingreasoningcapability}) when predicting a concatenated segmentation and glosses. We demonstrate this with a small pilot study using vanilla GRPO \citep{shao2024deepseekmathpushinglimitsmathematical} to optimize the \textsc{PolyGloss} ByT5 Concatenated model on Gitksan. We use $\beta=0.1$, $lr=5E-5$, batch and group sizes both 8, and train for 50 epochs. We use a temperature of 0.6, top-p of 0.9, and repetition penalty of 1.05 when sampling. 

We report results before and after tuning in \autoref{fig:grpo}, observing that all three metrics improve slightly. This approach could be scaled to the full training dataset as an additional post-training step, providing a method to improve the model without any additional labeled data. Furthermore, RL could be used when adapting the model to a new language that lacks gold-labeled morphological segmentations, since the alignment score will force the segmentation to (minimally) be aligned with the gold glosses.

\section{Related Work}
Research has explored automatic interlinear glossing using a variety of techniques including active learning \citep{palmer_computational_2010, palmer2009evaluating}, conditional random fields \citep{moeller_automatic_2018, mcmillan-major_automating_2020}, neural models \citep{moeller_automatic_2018, zhao-etal-2020-automatic}, and large language models \citep{ginn-etal-2024-teach, yang-etal-2024-multiple, elsner-liu-2025-prompt, shandilya-palmer-2025-boosting}. Our work is directly inspired by \citet{he-etal-2023-sigmorefun}, a submission to the 2023 SIGMORPHON Shared Task on Interlinear Glossing \citep{ginn-etal-2023-findings}, and \citet{ginn-etal-2024-glosslm}, both of which used multilingual pretraining on glossed text. 

Other work has focused on methods to automatically create IGT instances from other modalities, such as images of reference grammars \citep{round-etal-2020-automated}, LaTeX publications \citep{nordhoff-kramer-2022-imtvault}, and speech recordings \citep{he-etal-2024-wav2gloss}. Recently, \citet{aycockcan} and \citet{yang-etal-2025-linggym} proposed glossing as a method for testing LLMs' abilities to apply grammatical knowledge.

\section{Conclusion}
Tools for language documentation are only valuable when designed with the user---annotators and linguists---in mind. We address user feedback on automated glossing models and develop new multilingual models with major improvements over prior work. Most significantly, the \textsc{PolyGloss} models predict both morphological segments and interlinear glosses in a single forward pass, enabling more useful, interpretable, and trustworthy suggestions. We set a new state-of-the-art for glossing in several languages, as well as optimizing our model for segmentation accuracy and alignment between the two tasks. Finally, we offer practical recommendations for predicting performance and adapting the model to new data, keeping in mind computational constraints. 

Going forward, we plan to work with developers of annotation software such as ELAN \citep{ELAN2025} and FLEx \citep{Flex} to integrate our model into real-world documentation workflows. One straightforward solution is to use our model to predict segmentation and gloss lines for samples with transcribed text, filter predictions according to model confidence, and allow the user to accept, reject, or edit the predictions. 

\section*{Limitations}
We do not compare against closed-sourced LLMs for several reasons. First, the training datasets for these models are opaque, and with much of the test data being available online, there is risk of contamination (of course, this may also be true for the Qwen and Cohere models as well). Second, endangered language data often bears considerations of data sovereignty, and language communities often do not want their data to be sent to a third-party provider. Our models are open-source and open-weights and can be finetuned locally. 

For morphological segmentation, we do not differentiate between surface-level morphemes and underlying-form morphemes (and our dataset includes both). Thus, the task is not exactly identical across datasets, and there is no guarantee that the predicted segmentation may exactly match the input string.

We do not attempt to standardize the glosses in our dataset for two reasons. First, conventions and meanings vary greatly across annotators, and there is no way we could ensure the original intent of the annotator was preserved for the thousands of examples and languages in our dataset. Second, \citet{ginn-etal-2024-glosslm} tried standardizing glosses to the UniMorph conventions \citep{batsuren-etal-2022-unimorph} and found no evidence that it improved performance. Instead, our model outputs glosses according to the various conventions in its training data, and it can be adapted to a new convention via low-rank finetuning.

\section*{Ethical Considerations}
Automatic approaches to interlinear glossing are intended to help accelerate the language documentation process and contribute to the fight against language death. However, there is risk of misuse, and these systems should not fully replace human annotators, which could result in erroneous documentation that hinders downstream applications. All data was taken from existing work and used in accordance with the original stakeholders' wishes. Finally, our work used a large amount of computational resources, which inevitably bears an environmental cost.

\section*{Acknowledgments}
We thank Morris Alper for suggesting our "interleaved" format. Parts of this work were supported by the National Science
Foundation under Grant No. 2149404, "CAREER: From One Language to Another." This work also used  DeltaAI at NCSA through allocation CIS250116 from the Advanced Cyberinfrastructure Coordination Ecosystem: Services \& Support (ACCESS) program, which is supported by U.S. National Science Foundation grants \#2138259, \#2138286, \#2138307, \#2137603, and \#2138296.

Any opinions, findings, and conclusions or recommendations expressed in this material are those of the authors and do not necessarily reflect the views of the National
Science Foundation.

\bibliography{anthology-1,anthology-2,custom}

\appendix
\section{Base Model Ablation}
\label{sec:base_model_appendix}

In addition to the ByT5-based \textsc{PolyGloss} models, we also performed training on decoder-only instruction-tuned LLMs with significantly more parameters. We tried a number of hyperparameters through manually tuning, but were unable to match the ByT5 model performance. We report the best hyperparameters in \autoref{tab:hparams}.

\begin{table}[h]
    \small
    \centering
    \begin{tabular}{l|c c}
    \toprule
         & ByT5 & Qwen  \\
         \midrule
        LR & 5E-5 & 5E-5  \\
        Batch size & 64 & 18  \\
        Epochs & 15 & 15  \\
        \bottomrule
    \end{tabular}
    \caption{Hyperparameters for \textsc{PolyGloss} training.}
    \label{tab:hparams}
\end{table}



\begin{figure}[htb]
    \centering
    \includegraphics[width=\linewidth]{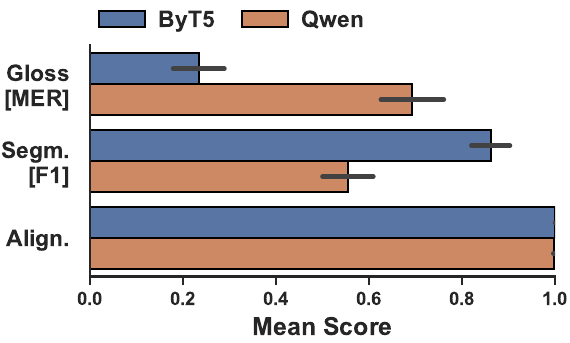}
    \caption{Scores for \textsc{PolyGloss} multilingual models using the Interleaved format and different base models. Scores are averaged across nine languages and reported with standard error.}
    \label{fig:ablation4}
\end{figure}

We provide average scores in \autoref{fig:ablation4} and full scores in \autoref{sec:ablation_numbers}. Generally, none of the other base models converged to a decent loss, and the scores are unsurprisingly far worse than ByT5. We hypothesize a few possible explanations. First, the decoder-only models use subword tokenizers (as opposed to the byte-level tokenizer of ByT5), which can cause issues for rare languages---particularly for segmentation, where the expected output is the input string with morpheme boundaries inserted, a very difficult task using multi-character subword tokens. Second, these models are trained on much more data (with instruction tuning and reinforcement learning) than ByT5, making it more difficult to escape the local minimum during continued finetuning. Third, these models are much larger, and our dataset's size relative to the parameter count may result in high-variance or uninformative gradients, impeding training. Still, we expect with the right hyperparameters, these models should be able to at least match the accuracy of the ByT5 model, which could be explored by future work.

\section{Ablation Results}
\label{sec:ablation_numbers}
Full results for the monolingual ablations and task format ablations are given in \autoref{tab:mer-glossing-ablations}, \autoref{tab:f1-segmentation-ablations}, and \autoref{tab:alignment-ablations}. We also provide evaluation metrics when evaluating \textsc{PolyGloss} only on long examples in \autoref{tab:long-ablation}.

\input{latex/tables/mer_glossing_abls}
\input{latex/tables/f1_segmentation_abls}
\input{latex/tables/alignment_abls}
\input{latex/tables/long_eval}

\section{In-Context Learning Failures}
\label{sec:appendix_llm_icl}
For the smaller Qwen model, the majority of outputs are malformed, resulting in very poor scores. The larger models conformed to the format consistently, but still produced poor outputs. We observe the several types of failure within the Tsez evaluation data:

\paragraph{Hallucinated Transcription}
The most common failure case is that the model produces a segmentation that does not match the input transcription. For the Qwen model, the predicted segmentation was often wildly divergent, such as the following example in which the model hallucinates a word and omits two words from the input:

\begin{small}
\begin{exe}
  \ex
  \gll Hič’č’a e\textgamma enä e\textcrlambda in: \\
       xan-a e\textcrlambda i-n \\
       khan-ERG say-PST.UNW
\end{exe}
\end{small}

For the larger models, this type of failure was also incredibly common, as in the following:

\begin{small}
\begin{exe}
  \ex
  \gll Nedur yeda ukaynosi , kidbä šebi roqä\textcrlambda in esirno . \\
       nedur-a yeda-q r-oq-n kid-a šebi r-oq-n-\textcrlambda in \\
       DEM1.IPL.OBL-ERG food-POSS.ESS IV-happen-PST.UNW girl-ERG what IV-happen-PST.UNW-QUOT
\end{exe}
\end{small}

Here, the segmentation is fairly close to the prediction transcription, but there are additional segments such as the first "-a" and "-q" inserted. In both cases, predicting an invalid segmentation naturally results in low-quality glosses. There is at least one character error in 79.9\% of examples with the Qwen model, 98.0\% of examples with Gemma, and 98.4\% of examples with Aya. We also compute the character error rate between the predicted and gold transcriptions (ignoring segmentation markers), with a CER of 0.66 for Qwen, 0.46 for Gemma, and 0.38 for Aya. This suggests that the smaller the model, the more the transcription diverges. We suspect these errors occur due to tokenization, which our \textsc{PolyGloss} model avoids by using byte-level tokenization.

\paragraph{Bad Format}
For the Qwen model, the output often does not follow the interleaved format, missing the interleaved parentheses used to indicate glosses. This occurs in 17.4\% of cases for Qwen, and never for the other two models. This could be addressed by trying different formats, providing more in-context examples, or using constrained decoding \citep{willard2023efficientguidedgenerationlarge} to force the correct format.

\paragraph{Repetition}
Finally, the LLM occasionally gets stuck in an endless repetition, in 2.5\% of cases for Qwen and never for the other models. This could potentially be addressed by tuning the repetition penalty.

\section{Use of AI Assistants}
We used AI assistants via GitHub's Copilot to review pull requests. We otherwise did not use AI assistance at any point of the research study.

\end{document}

%% file: latex/tables/mer_glossing_multiling.tex
\begin{table*}[!htb]
\centering
\resizebox{\textwidth}{!}{
    \def\arraystretch{1.2}
    \begin{tabular}{l | c c c c c c c c c | c}
    \toprule
    & arp & ddo & git & usp & ain & lez & ntu & nyb & ruc & 
    \textit{Avg.} \\
    \midrule
    Qwen 3 0.6B (ICL) & 0.868 & 0.904 & 0.919 & 0.730 & 0.773 & 0.895 & 0.877 & 0.706 & 0.883 & 0.839 \\
    Gemma 3 4B (ICL) & 0.489 & 0.597 & 0.826 & 0.476 & 0.351 & 0.668 & 0.473 & 0.430 & 0.723 & 0.559 \\
    Aya Expanse 8B (ICL) & 0.545 & 0.749 & 0.871 & 0.514 & 0.464 & 0.740 & 0.591 & 0.492 & 0.802 & 0.641 \\
     \textsc{GlossLM} & 0.161 & 0.095 & 0.870$^*$ & 0.163 & 0.909$^*$ & 0.940$^*$ & 0.893$^*$ & 0.990$^*$ & 0.731$^*$ & 0.639$^*$ \\
    \textsc{PolyGloss} (ByT5, multitask) & 0.177 & 0.089 & 0.603 & 0.162 & 0.122 & 0.383 & 0.189 & 0.328 & 0.329 & 0.265 \\
    \textsc{PolyGloss} (ByT5, interleaved) & \textbf{0.152} & \textbf{0.072} & \textbf{0.597} & \textbf{0.160} & \textbf{0.095} & \textbf{0.357} & \textbf{0.142} & \textbf{0.222} & \textbf{0.306} & \textbf{0.234} \\
    \bottomrule
    \end{tabular}%
}
\caption{Morpheme error rate ($\downarrow$) for \textbf{glossing} on the held-out test set with multilingual models. For \textsc{GlossLM}, the only eval languages explicitly included in the pretraining corpus are \texttt{arp}, \texttt{ddo}, and \texttt{git}, so scores on other languages (marked with $^*$) are very poor.} 
\label{tab:mer-glossing}
\end{table*}

%% file: latex/tables/f1_segmentation_multiling.tex
\begin{table*}[!htb]
\small
\centering
\resizebox{\textwidth}{!}{
    \def\arraystretch{1.2}
    \begin{tabular}{l | c c c c c c c c c | c}
    \toprule
    & arp & ddo & git & usp & ain & lez & ntu & nyb & ruc & 
    \textit{Avg.} \\
    \midrule
    Qwen 3 0.6B (ICL) & 0.091 & 0.096 & 0.050 & 0.297 & 0.233 & 0.118 & 0.133 & 0.298 & 0.190 & 0.167 \\
    Gemma 3 4B (ICL) & 0.505 & 0.310 & 0.159 & 0.508 & 0.575 & 0.338 & 0.435 & 0.617 & 0.345 & 0.421 \\
    Aya Expanse 8B (ICL) & 0.457 & 0.249 & 0.132 & 0.509 & 0.569 & 0.302 & 0.337 & 0.465 & 0.324 &  0.371\\
    \textsc{PolyGloss} (ByT5, multitask) & 0.903 & 0.967 & 0.605 & 0.850 & 0.963 & 0.826 & 0.925 & 0.938 & 0.763 & 0.860 \\
    \textsc{PolyGloss} (ByT5, interleaved) & \textbf{0.910} & \textbf{0.972} & \textbf{0.595} & \textbf{0.855} & \textbf{0.963} & \textbf{0.782} & \textbf{0.935} & \textbf{0.959} & \textbf{0.782} & \textbf{0.862} \\
    \bottomrule
    \end{tabular}%
}
\caption{Morpheme F1 ($\uparrow$) for \textbf{segmentation} on the held-out test set with multilingual models.} 
\label{tab:mer-segmentation}
\end{table*}

%% file: latex/tables/alignment_multiling.tex
\begin{table*}[!htb]
\small
\centering
\resizebox{\textwidth}{!}{
    \def\arraystretch{1.2}
    \begin{tabular}{l | c c c c c c c c c | c}
    \toprule
    & arp & ddo & git & usp & ain & lez & ntu & nyb & ruc & 
    \textit{Avg.} \\
    \midrule
    Qwen 0.6B (ICL) & 0.342 & 0.669 & 0.488 & 0.744 & 0.716 & 0.711 & 0.697 & 0.788 & 0.792 & 0.661 \\
    Gemma 3 4B (ICL) & 0.982 & 0.995 & 0.936 & 0.983 & 0.985 & 0.993 & 0.994 & 0.993 & 0.990 & 0.984 \\
    Aya Expanse 8B (ICL) & 0.957 & 0.953 & 0.956 & 0.973 & 0.979 & 0.882 & 0.978 & 0.985 & 0.986 & 0.961 \\
    \textsc{PolyGloss} (ByT5, multitask) & 0.984 & 0.995 & 0.919 & 0.990 & 0.982 & 0.973 & 0.985 & 0.988 & 0.941 & 0.973 \\
    \textsc{PolyGloss} (ByT5, interleaved) & \textbf{1.000} & \textbf{1.000} & \textbf{1.000} & \textbf{1.000} & \textbf{1.000} & \textbf{1.000} & \textbf{1.000} & \textbf{1.000} & \textbf{1.000} & \textbf{1.000} \\
    \bottomrule
    \end{tabular}%
}
\caption{\textbf{Alignment score ($\uparrow$)} between predicted segmentation and glosses on held-out test set, multilingual models.} 
\label{tab:alignment}
\end{table*}

%% file: latex/tables/mer_glossing_abls.tex
\begin{table*}[!htb]
\centering
\resizebox{\textwidth}{!}{
    \def\arraystretch{1.2}
    \begin{tabular}{l | c c c c c c c c c | c}
    \toprule
    & arp & ddo & git & usp & ain & lez & ntu & nyb & ruc & 
    \textit{Avg.} \\
    \midrule
    Finetuned ByT5 (separate) & \textbf{0.092} & \textbf{0.059} & 0.977 & \textbf{0.116} & 0.080 & 0.380 & 0.184 & 0.283 & \textbf{0.292} & 0.274 \\
    Finetuned ByT5 (joint) & 0.128  & 0.064 & 0.841 & 0.170 & \textbf{0.075} & 0.382 & 0.144 & 0.225 & 0.346 & 0.264 \\
    Pipeline & 0.140 & 0.076 & 0.972 & 0.173 & 0.096 & 0.454 & 0.238 & 0.314 & 0.433 & 0.322 \\
    \citet{girrbach-2023-tu-cl} & 0.185 & 0.110 & 0.732 & 0.242 & 0.353 & 0.392 & 0.266 & 0.246 & 0.512 & 0.338 \\
    \midrule
    \textsc{PolyGloss} (ByT5, multitask) & 0.177 & 0.089 & 0.603 & 0.162 & 0.122 & 0.383 & 0.189 & 0.328 & 0.329 & 0.265 \\
    \textsc{PolyGloss} (ByT5, concat) & 0.171 & 0.080 & \textbf{0.597} & 0.165 & 0.108 & \textbf{0.357} & 0.180 & 0.310 & 0.315 & 0.254 \\
    \textsc{PolyGloss} (ByT5, interleaved) & 0.152 & 0.072 & \textbf{0.597} & 0.160 & 0.095 & \textbf{0.357} & \textbf{0.142} & \textbf{0.222} & 0.306 & \textbf{0.234} \\
    \textsc{PolyGloss} (Qwen, interleaved) & 0.453 & 0.626 & 0.902 & 0.418 & 0.480 & 0.872 & 0.739 & 0.912 & 0.835 & 0.693 \\
    \bottomrule
    \end{tabular}%
}
\caption{Morpheme error rate ($\downarrow$) for \textbf{glossing} on the held-out test set. For \textsc{GlossLM}, the only languages explicitly included in the pretraining corpus are \texttt{arp}, \texttt{ddo}, and \texttt{git}, so scores on other languages (marked with $^*$) are very poor.} 
\label{tab:mer-glossing-ablations}
\end{table*}

%% file: latex/tables/f1_segmentation_abls.tex
\begin{table*}[!htb]
\small
\centering
\resizebox{\textwidth}{!}{
    \def\arraystretch{1.2}
    \begin{tabular}{l | c c c c c c c c c | c}
    \toprule
    & arp & ddo & git & usp & ain & lez & ntu & nyb & ruc & 
    \textit{Avg.} \\
    \midrule
    Finetuned ByT5 (separate) & 0.921 & 0.975 & 0.191 & 0.851 & 0.972 & 0.795 & 0.930 & 0.957 & 0.717 & 0.812 \\
    Finetuned ByT5 (joint) & 0.924 & 0.976 & 0.152 & 0.865 & 0.972 & 0.827 & 0.935 & 0.962 & 0.766 & 0.820  \\
    Pipeline & 0.921 & 0.975 & 0.191 & 0.851 & 0.972 & 0.795 & 0.930 & 0.957 & 0.717 & 0.812  \\
    \citet{girrbach-2023-tu-cl} & 0.531 & 0.447 & 0.241 & 0.548 & 0.348 & 0.349 & 0.441 & 0.730 & 0.494 & 0.459 \\
    Qwen 0.6B (ICL) &  & 0.096 &  & 0.297 & 0.233 & 0.118 & 0.133 & 0.298 & 0.190 &  \\
    Aya Expanse 8B (ICL) & 0.457 & 0.249 & 0.132 & 0.509 & 0.569 & 0.302 & 0.337 & 0.465 & 0.324 &  0.371\\
    \midrule
    \textsc{PolyGloss} (ByT5, multitask) & 0.903 & 0.967 & 0.605 & 0.850 & 0.963 & 0.826 & 0.925 & 0.938 & 0.763 & 0.860 \\
    \textsc{PolyGloss} (ByT5, concat) & 0.901 & 0.964 & 0.589 & 0.852 & 0.963 & 0.825 & 0.924 & 0.940 & 0.772 & 0.859 \\
    \textsc{PolyGloss} (ByT5, interleaved) & 0.910 & 0.972 & 0.595 & 0.855 & 0.963 & 0.782 & 0.935 & 0.959 & 0.782 & 0.862 \\
    \textsc{PolyGloss} (Qwen, interleaved) & 0.658 & 0.610 & 0.235 & 0.663 & 0.754 & 0.485 & 0.554 & 0.656 & 0.375 & 0.555 \\
    \bottomrule
    \end{tabular}%
}
\caption{Morpheme F1 ($\uparrow$) for \textbf{segmentation} on the held-out test set.} 
\label{tab:f1-segmentation-ablations}
\end{table*}

%% file: latex/tables/alignment_abls.tex
\begin{table*}[!htb]
\small
\centering
\resizebox{\textwidth}{!}{
    \def\arraystretch{1.2}
    \begin{tabular}{l | c c c c c c c c c | c}
    \toprule
    & arp & ddo & git & usp & ain & lez & ntu & nyb & ruc & 
    \textit{Avg.} \\
    \midrule
    Finetuned ByT5 (separate) & 0.980 & 0.988 & 0.650 & 0.977 & 0.979 & 0.885 & 0.907 & 0.963 & 0.904 & 0.915\\
    Finetuned ByT5 (joint) & 0.999 & \textbf{1.000} & 0.996 & \textbf{1.000} & \textbf{1.000} & \textbf{1.000} & \textbf{1.000} & \textbf{1.000} & 0.997 & 0.999 \\
    Pipeline & 0.998 & 0.989 & 0.464 & \textbf{1.000} & 0.994 & 0.522 & 0.785 & 0.914 & 0.713 & 0.820 \\
    \citet{girrbach-2023-tu-cl} & 0.998 & 0.998 & \textbf{1.000} & 1.000 & 0.895 & 0.985 & 0.976 & 0.994 & 0.941 & 0.976 \\
    \midrule
    \textsc{PolyGloss} (ByT5, multitask) & 0.984 & 0.995 & 0.919 & 0.990 & 0.982 & 0.973 & 0.985 & 0.988 & 0.941 & 0.973 \\
    \textsc{PolyGloss} (ByT5, concat) & 0.996 & 0.997 & 0.965 & \textbf{1.000} & 0.994 & 0.986 & 0.995 & 0.999 & 0.973 & 0.989 \\
    \textsc{PolyGloss} (ByT5, interleaved) & \textbf{1.000} & \textbf{1.000} & \textbf{1.000} & \textbf{1.000} & \textbf{1.000} & \textbf{1.000} & \textbf{1.000} & \textbf{1.000} & \textbf{1.000} & \textbf{1.000} \\
    \textsc{PolyGloss} (Qwen, interleaved) & 0.999 & 0.999 & 0.996 & 0.999 & 1.000 & 0.988 & 1.000 & 0.997 & 1.000 & 0.997 \\
    \bottomrule
    \end{tabular}%
}
\caption{\textbf{Alignment score ($\uparrow$)} between morphological segmentation and predicted glosses on the held-out test set.} 
\label{tab:alignment-ablations}
\end{table*}

%% file: latex/tables/long_eval.tex
\begin{table*}[!htb]
\small
\centering
\resizebox{\textwidth}{!}{
    \def\arraystretch{1.2}
    \begin{tabular}{l | c c c c c c c c c | c}
    \toprule
    & arp & ddo & git & usp & ain & lez & ntu & nyb & ruc & 
    \textit{Avg.} \\
    \midrule
    \textit{Glossing} & \\
        \textsc{PolyGloss} (Qwen, interleaved) & 0.453 & 0.626 & 0.902 & 0.418 & 0.480 & 0.872 & 0.739 & 0.912 & 0.835 & 0.693 \\
    \textsc{PolyGloss} (ByT5, multitask, long only) & 0.167 & 0.106 & 0.613 & 0.159 & 0.129 & 0.412 & 0.186 & 0.300 & 0.332 & 0.267 \\
    \textsc{PolyGloss} (ByT5, concat, long only) & 0.161 & 0.099 & 0.628 & 0.169 & 0.115 & 0.409 & 0.191 & 0.293 & 0.320 & 0.265 \\
    \textsc{PolyGloss} (ByT5, interleaved, long only) & 0.146 & 0.089 & 0.603 & 0.151 & 0.102 & 0.317 & 0.129 & 0.206 & 0.294 & 0.226 \\
    \midrule
    \textit{Segmentation} & \\
    \textsc{PolyGloss} (ByT5, multitask, long only) & 0.919 & 0.959 & 0.620 & 0.829 & 0.967 & 0.816 & 0.919 & 0.945 & 0.800 & 0.864 \\
    \textsc{PolyGloss} (ByT5, concat, long only) & 0.917 & 0.958 & 0.573 & 0.837 & 0.969 & 0.816 & 0.921 & 0.948 & 0.808 & 0.861 \\
    \textsc{PolyGloss} (ByT5, interleaved, long only) & 0.924 & 0.965 & 0.628 & 0.849 & 0.961 & 0.791 & 0.943 & 0.966 & 0.825 & 0.872 \\
    \midrule
    \textit{Alignment} & \\
    \textsc{PolyGloss} (ByT5, multitask, long only) & 0.987 & 0.995 & 0.903 & 0.985 & 0.981 & 0.972 & 0.975 & 0.992 & 0.951 & 0.971 \\
    \textsc{PolyGloss} (ByT5, concat, long only) & 0.996 & 0.997 & 0.932 & 0.999 & 0.991 & 0.989 & 0.989 & 1.000 & 0.981 & 0.986 \\
    \textsc{PolyGloss} (ByT5, interleaved, long only) & 1.000 & 1.000 & 1.000 & 1.000 & 1.000 & 1.000 & 1.000 & 1.000 & 1.000 & 1.000 \\
    \bottomrule
    \end{tabular}%
}
\caption{Glossing, segmentation, and alignment scores for \textsc{PolyGloss} models only evaluated on very long inputs (75th percentile in length per language).} 
\label{tab:long-ablation}
\end{table*}